\documentclass{l4dc2025}
\usepackage{dsfont}
\usepackage{booktabs}
\usepackage{caption}
\usepackage{subcaption}
\usepackage{enumitem}
\usepackage{amsmath}
\usepackage{threeparttable}
\usepackage{wrapfig}
\usepackage{algorithm,algcompatible}
\algnewcommand\INPUT{\item[\textbf{Input:}]}%
\algnewcommand\OUTPUT{\item[\textbf{Output:}]}%
\newtheorem{mydef}{Definition}
\newtheorem{mythm}{Theorem}
\newtheorem{myprob}{Problem}

\newtheorem{mypro}{Proposition}

\DeclareMathOperator*{\argmin}{arg\,min}

\newcommand{\reals}{\mathbb{R}}
\newcommand{\preals}{\reals_{\geq 0}}
\newcommand{\nat}{\mathbb{N}}

\newcommand{\haut}{H}

\newcommand{\odesol}{\varphi}
\newcommand{\initcond}{x}

\newcommand{\safesw}{\mathcal{S}}
\newcommand{\unsafesw}{\mathcal{U}}

\newcommand{\BV}{B}

\newcommand{\Cc}{\mathcal{C}}
\newcommand{\BVg}{\BV_{\gamma}}

\newcommand{\param}{\theta}

\newcommand{\loss}{h}

\newcommand{\revise}[1]{\textcolor{black}{#1}}

\title[Learn Local CBFs for Hybrid Systems]{Learning Local Control Barrier Functions for Hybrid Systems}
\usepackage{times}

\author{%
 \Name{Shuo Yang}$^{*}$$^{1}$ \Email{yangs1@seas.upenn.edu}
 \AND
 \Name{Yu Chen}$^{*}$$^{2}$ \Email{yuchen26@sjtu.edu.cn}
 \AND
 \Name{Xiang Yin}$^{2}$ \Email{yinxiang@sjtu.edu.cn}
 \AND
 \Name{George J. Pappas}$^{1}$ 
\Email{pappasg@seas.upenn.edu}
\AND
 \Name{Rahul Mangharam}$^{1}$ \Email{rahulm@seas.upenn.edu}\\
 \addr $^{1}$University of Pennsylvania, Philadelphia, PA, USA \\
 \addr $^{2}$Shanghai Jiao Tong University, Shanghai, China \\
 \addr $^{*}$ Indicates equal contribution%
}

\begin{document}

\maketitle

\begin{abstract}%
 Hybrid dynamical systems are ubiquitous as practical robotic applications often involve both continuous states and discrete switchings.
 Safety is a primary concern for hybrid robotic systems.
 Existing safety-critical control approaches for hybrid systems are either computationally inefficient, detrimental to system performance, or limited to small-scale systems.
 To amend these drawbacks, in this paper, we propose a learning-enabled approach to construct local Control Barrier Functions (CBFs) to guarantee the safety of a wide class of nonlinear hybrid dynamical systems.
 The end result is a safe neural CBF-based switching controller.
 Our approach is computationally efficient, minimally invasive to any reference controller, and applicable to large-scale systems.
 We empirically evaluate our framework and demonstrate its efficacy and flexibility through two robotic examples including a high-dimensional autonomous racing case, against other CBF-based approaches and model predictive control.
\end{abstract}


\setlength{\textfloatsep}{1pt}
\section{Introduction}

Consider the following safety-critical scenarios: 
1) automated vehicles switching from eco driving mode to sport driving mode; 
2) bipedal robots walking in the warehouse to assist human; and 
3)  autonomous drones flying from high-pressure area to low-pressure area.
The common feature of these ubiquitous autonomous systems is that they are all \emph{hybrid dynamical systems}, which involve both continuous state evolution and discrete mode switching. 
Such discrete mode switching can be actively triggered by high-level logical decision-making or passively induced by sudden changes in the underlying physical environments.
 
Safety-critical control is a fundamental problem for hybrid dynamical systems.
In the past decades, many approaches have been proposed to guarantee safety for hybrid systems such as model predictive control (MPC)~\cite{borrelli2017predictive} and Hamilton-Jacobi reachability (HJ-reach)~\cite{tomlin2003computational, bansal2017hamilton}. 
MPC provides safety guarantees by encoding safety constraints in a receding horizon optimization problem, which suffers from high online computation cost, especially when the dynamics \revise{are} nonlinear or the horizon is large.
HJ-reach methods  compute optimal safe controllers by solving \revise{partial differential equations (PDEs)} through dynamic programming, which are difficult to \revise{apply} to high-dimensional systems. 
Other approaches such as computing controlled invariant sets~\cite{legat2018computing} are also difficult to apply to general nonlinear systems.

Control barrier \revise{functions (CBFs)}~\cite{ames2016control, ames2019control, hsu2023safety} \revise{have recently emerged} as a promising approach \revise{to safety-critical control} since \revise{they} can explicitly specify a safe control set and enforce the invariance of safe sets.
This is achieved by constructing a CBF-based safety filter that projects unsafe reference control into the safe control set.
Compared with MPC, CBF enjoys low \revise{online} computation cost and can be applied to any \revise{real-time} reference controller as a flexible safety filter.
\revise{The offline computed safe control policy by HJ-reach is generally overly conservative when applied, but the CBF constraint can be incorporated in an online Quadratic Program based minimally invasive controller such that the controlled behavior is not restrictive~\cite{choi2021robust}.}
In the context of safety control for hybrid systems, \cite{lindemann2021learning, robey2021learning} propose to use a common CBF shared by all modes to ensure global safety. 
However, such a common/global CBF may be overall conservative to degrade the system performance. 

In our recent work~\cite{yang2023safe}, we propose a local CBFs-based approach to address the safety control problem for hybrid systems. 
Specifically, we assume that, for each dynamical mode of the hybrid system, there already exists a CBF that can ensure safety locally within this model.
However, it is discovered that local safety for each mode is not sufficient to ensure safety globally as unsafe behaviors may occur under discrete jumps. 
To address this issue,  safe switching sets are proposed to further refine the original local CBFs, by taking discrete jumps into account, such that the refined CBF-based controller can guarantee global safety.
It is demonstrated that local CBFs method is more general than global CBF and enjoys better control performance.

Although the local CBF-based approach provides a promising framework for ensuring global safety for hybrid systems, the original approach of~\cite{yang2023safe} has \revise{a} significant limitation as it is only applicable to small-scale systems (the state dimension cannot be greater than 5 practically) because HJ-reach is used in the local CBF refinement.
To overcome this issue,  built upon the framework of \cite{yang2023safe}, 
we propose a learning-based local CBF refinement approach in this paper, which can scale to high-\revise{dimensional} systems. Specifically, we inherit the spirit of reducing refining local CBF to solving HJ-reach PDE. However, we do not resort to traditional approach to solve HJ-reach PDE, i.e., discretizing the state space into grids and solve it through dynamic programming.
Instead, we leverage recent advances in neural PDE solvers~\cite{bansal2021deepreach, sirignano2018dgm, han2018solving, berg2018unified} to approximate the solution of HJ-reach PDE through a deep neural network (DNN). This leads to a DNN representation of our refined CBF, and based on which a switching neural local CBFs-based controller can be obtained to guarantee global safety.
The contributions of our work are as follows:

\begin{itemize}[itemsep=2pt,topsep=0pt,parsep=0pt]
    \item 
    We propose a learning-based approach to refine local CBFs for hybrid systems to ensure safety globally. 
    Therefore, by leveraging the power of deep learning, 
    our approach is applicable to high-dimensional systems as explicit dynamic programming can be avoided;
    \item 
    Particularly, we  prove that computing \emph{backward unsafe set} for \emph{unsafe switching set} using HJ-reach in~\cite{yang2023safe} is not necessary. 
    This further simplifies the safety controller synthesis procedure in original local CBF framework;
    \item 
    Furthermore, 
    experimental simulations are provided to illustrate the benefits of our approach against other CBF-based approaches and MPC. 
    In particular, we apply our approach to a high-dimensional autonomous racing scenario, for which the original local CBF approach cannot handle due to the curse of dimensionality.
    Our code is publicly available\footnote{\url{https://github.com/shuoyang2000/neural_hybrid_cbf}}.
\end{itemize}

\subsection{Related Work}


\emph{Learning Control Barrier Function: }
An open problem for CBF is how to construct a valid CBF through a systematic approach.
This problem is challenging as the CBF constraint should be satisfied for any safe state for a nonlinear dynamic with input constraints.
One recent promising line of work tackling this challenge is to learn a CBF from data.
These techniques either learn a CBF of a given system~\cite{srinivasan2020synthesis, robey2020learning, abate2021fossil} or simultaneously learn a control policy with CBF safety filter~\cite{qin2021learning, wang2023enforcing}.
Unlike these existing works which mostly cast the learning CBF problem into supervised learning given safe and unsafe datasets, we reduce the CBF construction into a HJ-reach problem~\cite{choi2021robust} and use DNN to approximate the solution of HJ-reach PDE by self-supervision.

\emph{Safety Control for Hybrid Systems: }
Various safety verification and control synthesis techniques for hybrid systems are studied in the past decades; see, e.g.,~\cite{mhaskar2005robust, benerecetti2013automatic, ivanov2019verisig, phan2019neural}.
Barrier function was introduced in~\cite{prajna2004safety} to verify the safety of hybrid systems and has been studied widely~\cite{glotfelter2019hybrid, maghenem2019characterizations, nejati2022compositional} since it can provide provable safety guarantees.
In~\cite{lindemann2021learning, robey2021learning}, hybrid CBF is proposed as a principled control method to synthesize safe controllers for hybrid systems. Furthermore, local CBFs-based method emerges and is demonstrated to be effective and general in the recent work~\cite{yang2023safe}.
In this paper, we propose a learning-based approach in local CBFs framework to enable the applicability to large-scale hybrid systems.

\vspace{-3mm}
\section{Preliminary and Problem Formulation}
We denote by $\mathbb{R}$ and $\mathbb{R}^n$ the sets of real numbers and real $n$-dimensional vectors, respectively. 
Given a set $X$, we denote by $\mathcal{P}(X)$ its powerset and $\sup X$ its supremum element.
Set $\nat$ denotes all natural numbers including zero.
Function $\alpha$: $\mathbb{R}\rightarrow \mathbb{R}$ denotes an extended class $\mathcal{K}_{\infty}$ function, i.e., a strictly increasing function with $\alpha(0)=0$.
\vspace{-3mm}
\subsection{Control Barrier Functions}
We consider a continuous control-affine system
\vspace{-3mm}
\begin{equation}\label{eq:system}
    \Dot{x} = F(x, u) = f(x) + g(x)u, \quad x(0)=x_0,
    \vspace{-3mm}
\end{equation}
where $x(t)\in\mathbb{R}^n$ is  the system state  at time $t\in\mathbb{R}_{\ge 0}$, and functions $f$ and $g$ are assumed to be locally Lipschitz.
We denote the trajectory (solution) of system~(\ref{eq:system}) at time $t$ starting at $x_0, t_0$ under control signal $u$ by $\xi^{F, u}_{x_0, t_0}(t)$. 
Safety is framed in the context of enforcing set invariance in the state space through appropriate control law $u$.
Consider a continuously differentiable function $h: D \rightarrow \mathbb{R}$ where $D\subset \mathbb{R}^n$.
The safe set $\mathcal{C}$ that we aim to render forward invariant is represented by the super-level set of $h$, i.e., $\mathcal{C}=\{x\in D:h(x)\ge 0\}$.

We say $h$ is a \emph{control barrier function} (CBF) if there exists an extended class $\mathcal{K}_{\infty}$ function $\alpha(\cdot)$ such that, for  system (\ref{eq:system}),  for all $x\in D$, it holds that:
$\sup_{u \in U}\left[\frac{\partial h(x)}{\partial x}\big(f(x)+g(x)u\big)\right]\ge -\alpha(h(x)),$ 
where $U\subset \mathbb{R}^m$ is the admissible control set.
Given the CBF $h(x)$, the set of all control values that render $\mathcal{C}$ safe is thus given by 
$
K_{cbf}(x):=\left\{u \!\in \!U: \frac{\partial h(x)}{\partial x}\big(f(x)+g(x)u\big)\!\ge\! -\alpha(h(x))\right\}$,
which we denote as the safe control set.
The next theorem shows that the existence of a CBF implies  safety:
\begin{mythm}\hspace{-0.3pt}\cite{ames2016control}
\label{thm:cbf}
Assume $h(x)$ is a CBF on $D \supset \mathcal{C}$. Then any Lipschitz continuous controller $u(x)$ such that $u(x) \in K_{cbf}(x)$ for all $x \in \mathcal{C}$ will render the set $\mathcal{C}$ forward invariant.
\end{mythm}
Typically, CBF works as a safety filter and a safe controller is obtained by solving the quadratic programming problem~\cite{ames2019control}:
$u(x)= \argmin_{u\in K_{cbf}(x)} \lVert u - k(x) \rVert^2$,
which is referred as CBF-QP and $k(x)$ is any given potentially unsafe reference controller 
and the resulting $u(x)$ is the safe actual control input applied for state $x$.
\vspace{-3mm}
\subsection{Hybrid Systems}  
We model hybrid systems with both continuous dynamic flows and discrete dynamic transitions following similar formalism of~\cite{goebel2009hybrid, yang2023safe}:
\begin{mydef}[Hybrid Systems]\label{def:hybrid_system}
  A hybrid system $\haut$ is a tuple 
$\haut =\langle X, Q, U, F, 
\texttt{Guard}\rangle$, where $X\subseteq D\subseteq\mathbb{R}^n$ and $Q$ are the continuous flow state space and discrete operation modes set respectively, $U\subseteq \mathbb{R}^m$ is the admissible control input set, $F: Q\times X\times U \rightarrow X$ is the continuous-time dynamic flow, and $\texttt{Guard} : Q \times Q \rightarrow \mathcal{P}(X)$ denotes the guard set that triggers mode switching.
\end{mydef}
We consider the $\texttt{Reset map}: Q\times Q\times X \rightarrow X$ as the identify function, i.e., the continuous system state remains the same for any mode jump: $\texttt{Reset}(q, q', x)=x$., so we are essentially considering \emph{switched systems} in this work.
We define $F$ as a control affine system with admissible control set $U_q\subseteq U$ for mode $q$:

\vspace{-4mm}
\begin{equation}
\label{eq:aff_dyn}
    \Dot{x}=F_q(x, u)=f_q(x)+g_q(x)u, \quad u\in U_q.
    \vspace{-2mm}
\end{equation}
\revise{For a given switching feedback control law $u: Q \times X\rightarrow U$,}
a solution (trajectory) to  hybrid system $H$ is a sequence $(q_i, \odesol_i, \delta_i)_{i\in N}$, where 
$N$ is $\nat$ or a bounded subset of $\nat$, $q_i \in Q$ is the discrete mode, $\odesol_i : X \times \preals \rightarrow X$ is the continuous state evolution, and $\delta_i$ is the duration of mode $q_i$ (i.e., dwell time).
The switching time from mode $q_{i}$ to $q_{i+1}$ is denoted by 
$\tau_{i+1} = \sum_{j=0}^{i} \delta_j, \forall i\in N \backslash\sup N.$
The solution is formally defined below.
\begin{mydef}[Hybrid System Solution]\label{def:hybrid_solutions}
Given hybrid system $\haut$ with a set of initial conditions $Q_0 \times X_0\subseteq Q \times X$, \revise{given switching feedback control law $u$,} a solution of $\haut$ is a sequence $(q_i, \odesol_i, \tau_i)_{i\in N}$ such that
 \begin{enumerate}[itemsep=2pt,topsep=0pt,parsep=0pt]
\item[(i)]
$(q_0, \initcond_0)\in Q_0 \times X_0$ is the initial state at time $\tau_0 = 0$.
\item[(ii)] 
\label{def:hsol:odesol}
$\forall i\in N$ with $i>0$  and  $t \in [\tau_{i}, \tau_{i+1}]$, $\odesol_i(\initcond_i,t)$ is the solution of (\ref{eq:aff_dyn}) for mode $q_i$ \revise{under control $u(q_i, \cdot)$} with initial condition $\initcond_i = \odesol_{i-1}(\initcond_{i-1},\tau_{i})$.  If $\tau_{i+1}=\infty$,   $t$ ranges over $[\tau_i,\tau_{i+1})$.
\item[(iii)]
$\forall i\in N$ with $i>0$, if $\tau_{i}<\infty$, then 
$\odesol_{i-1}(x_{i-1},\tau_{i})\in \texttt{Guard}(q_{i-1}, q_{i}).$
\end{enumerate}
\end{mydef}
We denote by $\mathcal{B}_{\haut}$ the set of all solutions of $\haut$ and define the set of all possible mode transition pairs of $\haut$ as
$
    \mathcal{T}(\haut) = \{(q_i, q_{i+1})_{i \in N\backslash \sup N}:
(q_i,\odesol_i,\delta_i)_{i \in N}\!\in\!\mathcal{B}_{\haut}\}.
$
\vspace{-3mm}
\subsection{Problem Formulation}
We recall the notions of  transition safety and global safety for hybrid systems~\cite{yang2023safe}.

\begin{mydef}[Transition Safety]
Given a hybrid system $\haut$, a pair of modes $(q,q') \in Q\times Q$, and safe sets $\mathcal{C}_q, \mathcal{C}_{q'}\subset D$ for modes $q$ and $q'$ respectively,  
we say that $\haut$ is \emph{$(q, q')$-safe} w.r.t.\ $\mathcal{C}_q$ and $\mathcal{C}_{q'}$ 
    if for any initial state $(q_0,\initcond_0)$ with $q_0=q$ and $\initcond_0\in\Cc_q$, any trajectory 
    $(q_i,\odesol_i,\delta_i)_{i\in\{0,1\}}$ of $\haut$ with $q_1 = q'$ satisfies:
    1) $\odesol_0(\initcond_0,t) \in \mathcal{C}_q,\forall t \in [\tau_0, \tau_1]$; and 2) $\odesol_1(\initcond_1,t) \in \mathcal{C}_{q'},\forall t \in [\tau_1,\tau']$, where 
        $\tau'=\tau_2$ if $\delta_1<\infty$.
        \revise{When $\delta_1=\infty$, $t$ ranges over $[\tau_1, \infty)$.}
\end{mydef}
\vspace{-4mm}
\begin{mydef}[Global Safety]
\label{def:global:multi:cbf} 
Given a hybrid system $\haut$,  safe sets $\mathcal{C}_q\subset D$ for any mode $q\in Q$, we say that $\haut$ is \emph{globally safe} w.r.t.\ $\{\mathcal{C}_q\}_{q\in Q}$ if $\haut$ is $(q, q')$-safe for any $(q, q')\in \mathcal{T}(\haut)$.
\end{mydef}
Intuitively, $(q, q')$-safety concerns the local transition safety from modes $q$ to   $q'$, and the system is globally safe if all trajectories are safe for all possible transition mode pairs.
As we stated in the introduction, we assume that a local CBF can be found to ensure safety within each mode, 
and our main focus is to ensure safety for mode switchings. 
We formulate the problem as follows.
\begin{myprob}\label{problem}
For any mode $q\in Q$ of $\haut$, assume that there exists a local CBF $h_q$ for 
    (\ref{eq:aff_dyn})
    with the corresponding safe set $\mathcal{C}_q=\{x\in D\subset\mathbb{R}^n: h_q(x)\ge 0\}$.
Find switching control laws that guarantee transition safety  and global safety of $\haut$.
\end{myprob}

\vspace{-6mm}
\section{Methodology: Learning Local CBFs}
In this section, we present our main methodology for \revise{solving} the safety control problem. 
Our approach builds upon the framework of local CBF as proposed in~\cite{yang2023safe}. However, rather than explicitly solving this problem, we introduce a novel, highly efficient learning-based approach to address the challenge of the curse of dimensionality in high-dimensional systems.
\vspace{-3mm}
\subsection{Local CBFs for Safety Control}\label{sec: review_safety_control}
 
Let $q,q'\in Q$ be two modes. 
Then the   \emph{safe switching set} and the \emph{unsafe switching set} for mode jump $q\rightarrow q'$ of $\haut$ are respectively defined by 
\vspace{-3mm}
\begin{equation}
\safesw_{q, q'}=\texttt{Guard}(q, q')\cap\mathcal{C}_q\cap\mathcal{C}_{q'},\quad 
\unsafesw_{q, q'}=(\texttt{Guard}(q, q')\cap\mathcal{C}_q)\setminus \safesw_{q, q'}.
\vspace{-3mm}
\end{equation} 
Furthermore, we define the \emph{unsafe backward set} $q\rightarrow q'$ in $\haut$ 
as the set of states from which unsafe switching is not avoidable, i.e.,  
\vspace{-3mm}
\begin{equation}
\texttt{BackUnsafe}_{q, q'}= 
         \{x_0 \in C_q  : \forall u(\cdot) \in U_{[0, \infty)}, \exists T \in \preals, 
        \text{s.t. } \xi^{F_q, u}_{x_0, t_0}(T) \in \unsafesw_{q, q'} \}.
    \vspace{-3mm}
    \end{equation} 
The procedure of synthesizing (transition) safety controller for hybrid system $\haut$ consists of 4 steps:

    \begin{itemize}[itemsep=2pt,topsep=0pt,parsep=0pt]
\item 
\textbf{Step 1:} 
Identify the safe switching set $\safesw_{q, q'}$ and the unsafe switching set $\unsafesw_{q, q'}$ for each $q\rightarrow q'$.  
\item 
\textbf{Step 2:}  
For each $q\rightarrow q'$, based on $\unsafesw_{q, q'}$, compute  the backward reachable set
$\texttt{BackUnsafe}_{q, q'}$.  
\item 
\textbf{Step 3:}
Refine the original local CBFs $\{\mathcal{C}_q\}_{q\in Q}$ by considering the computed new unsafe set through dynamic programming-based techniques~\cite{tonkens2022refining}.
\item 
\textbf{Step 4:}
Use the refined CBFs to control the hybrid systems and transition safety is ensured.
\end{itemize}


The above procedure can provably ensure the system safety.
However, both the backward reachable set computation (Step 2) and CBF refinement process (Step 3) rely on Hamilton-Jacobi reachability, which is limited to small-scale systems.
This is because HJ-reach involves solving a PDE through dynamic programming, whose computational complexity scales exponentially w.r.t. the system state dimension.
Practically, it can only handle low-dimensional models with state dimension less than 6.
To overcome the ``curse of dimensionality'', we propose a learning-based method instead in this work to synthesize safety controllers, which is introduced in the next subsection.
\vspace{-3mm}
\subsection{Learn to Refine Local CBFs}

We first introduce several necessary notions to facilitate the presentation of our method.
We consider a safety constraint set $\mathcal{L}=\{x\in X: \ell(x)\ge 0\}$ that is the super-level set of a continuous function $\ell:X\rightarrow \mathbb{R}$.
However, not every state in $\mathcal{L}$ can be guaranteed safe in a given time horizon considering system dynamics and input constraints.
Thus, for system~(\ref{eq:system}), we define the viability kernel\footnote{Note that the literature (e.g.,~\cite{bansal2017hamilton, choi2021robust}) define the viability kernel and related concepts on a negative time interval $[t, 0]$ where $t<0$. We define them on positive time intervals instead to be consistent with hybrid systems definition.} $\mathcal{S}(t)$ as the largest (time-varying) control invariant subset of $\mathcal{L}$ within duration $[0, t]$:
\vspace{-2mm}
\begin{equation}
        \mathcal{S}(t):= \{x \in \mathcal{L}: \exists u(\cdot) \in U_{[0, t]} \text{ s.t. } \xi_{x,0}^{F, u}(s) \in \mathcal{L},\forall s \in [0, t]\}. \nonumber
        \vspace{-2mm}
\end{equation}
Intuitively $\mathcal{S}(t)\subseteq \mathcal{L}$ contains all initial states from which there exists a control signal that keeps  system~(\ref{eq:system}) stay inside the safe constraint set $\mathcal{L}$ over a time duration $[0, t]$.
When $t\rightarrow\infty$, we know that every state in \revise{$\mathcal{S}_{\infty}:=\text{lim}_{t\rightarrow \infty}\mathcal{S}(t)$} can stay inside forever.
Note that the CBF safe set $\mathcal{C}$ is a subset of \revise{$\mathcal{S}_{\infty}$}, i.e.,  $\mathcal{C}\subseteq \revise{\mathcal{S}_{\infty}}$.
Now we define a special CBF called Control Barrier-Value Function (CBVF) that has both state and time as inputs.
\vspace{-2mm}
\begin{mydef}(Control Barrier-Value Function~\cite{choi2021robust})
    A Control Barrier-Value Function $B_\gamma: X \times [0, \infty) \to \mathbb{R}$ satisfies:
\vspace{-2mm}\begin{equation}\label{eq:cbvf}
     \BVg(x,t) := \max_{u(\cdot)\in U_{[0, t]}} \min_{s\in [0, t]} e^{\gamma s} \ell(\xi_{x, 0}^{F, u}(s)),
     \vspace{-2mm}
    \end{equation}
    for some $\gamma \in \mathbb{R}_{\geq 0}$ and $\forall t\geq 0$, with initial condition $B_\gamma(x, 0) = \ell(x)$.
\end{mydef}
\vspace{-2mm}
The main property of the CBVF is that it recovers the viability kernel~\cite{choi2021robust}, i.e., 
$\forall t \geq 0$, $\gamma \in \mathbb{R}_{\geq 0}$, $\mathcal{C}_{B_\gamma}(t) = \mathcal{S}(t)$, where $\mathcal{C}_{B_\gamma}(t) = \{x\in D\subset\mathbb{R}^n: \BVg (x,t) \geq 0\}$.
The CBVF in~(\ref{eq:cbvf}) can be computed using dynamic programming, which results in the following CBVF Hamilton-Jacobi-Isaacs Variational Inequality (HJI-VI) \revise{with boundary condition $B_\gamma(x, 0) = \ell(x)$}:
\vspace{-2mm}
\begin{equation}\label{pde-cbvf}
    \min\left\{-\frac{\partial B_{\gamma}(x, t)}{\partial t} + \textsf{Ham}(x, t), \ell(x) - B_{\gamma}(x, t)\right\}=0,
    \vspace{-2mm}
\end{equation}
where $\textsf{Ham}$ is the Hamiltonian which optimizes over
the inner product between the spatial gradients of the $B_{\gamma}$ and the flow field of the dynamics:
$\textsf{Ham}(x, t) = \max_{u\in U} \langle \frac{\partial B_{\gamma}(x, t)}{\partial x} , F(x, u)\rangle + \gamma B_{\gamma}(x, t).$

The CBVF is constructed during the CBF refinement (Step 3 of Section~\ref{sec: review_safety_control}) by using a safety constraint set $\mathcal{L}$ excluding backward unsafe set.
The obtained CBVF serves as the new CBF that ensures the safety of hybrid systems.
In practice, to construct CBVF, we should solve the HJI-VI~(\ref{pde-cbvf}) on a spatially discretized grid representing the continuous state space, which results in exponential computational complexity w.r.t. the system state dimensionality and thus is intractable for large systems.
To handle this challenge, we propose to use \revise{a} learning-based approach to approximate the solution of the HJ PDE without explicitly solving it.
To avoid HJ computation, we first prove that the $\texttt{BackUnsafe}$\footnote{\revise{Since we mostly discuss how to ensure transition safety in this work, two modes switching is concerned in most cases. Thus,} we omit the subscript $(q, q')$ if the context is clear, so does the unsafe switching set $\unsafesw$.} set computation can be avoided and then present our learning-based solution.
All proof is delayed in the Appendix.

\begin{mythm}\label{thm:no-backunsafe}
    The viability kernel $\revise{\mathcal{S}_{\infty, b}}$ of $\mathcal{L}\setminus\texttt{BackUnsafe}$ equals to viability kernel $\revise{\mathcal{S}_{\infty, u}}$ of $\mathcal{L}\setminus\unsafesw$.
\end{mythm}
Since CBVF recovers the largest viability kernel, we immediately get following result.
\begin{mypro}\label{prop:same-cbvf}
    The CBVF $B^{b}_{\gamma}(x, \infty)$ constructed based on the constraint set $\mathcal{L}\setminus\texttt{BackUnsafe}$ shares the same safe set with the CBVF $B^{u}_{\gamma}(x, \infty)$ constructed based on the constraint set $\mathcal{L}\setminus\unsafesw$, i.e., $\mathcal{C}_{B^b_\gamma}(\infty)=\mathcal{C}_{B^u_\gamma}(\infty)$.
\end{mypro}
Based on Proposition~\ref{prop:same-cbvf}, we know that it is not necessary to compute the $\texttt{BackUnsafe}$ set as the constructed CBVF will have the same safe set. 
This can help us avoid the Step 2 in Section~\ref{sec: review_safety_control}.
Now, we \revise{introduce} a deep learning-based method to obtain the CBVF $B(x, t)$\footnote{For the simplification of notation, we use $B(x, t)$ to denote $B_{\gamma}(x, t)$ if there is no ambiguity in the context.} for the new safety constraint set $\mathcal{L}\setminus\unsafesw$.
We represent $\unsafesw$ as the zero-level set of a function $\ell_\unsafesw:X\rightarrow \mathbb{R}$, i.e., $\unsafesw=\{x\in X: \ell_\unsafesw(x)\le 0\}$, so we can represent the new safety constraint set as $\mathcal{L}\setminus\unsafesw=\{x\in X: \ell_{new}(x)\ge 0\}$, where $\ell_{new}(x)=\min(\ell_\unsafesw(x), \ell(x))$.

The training procedure is summarized in Algorithm~(\ref{alg:train}).
Specifically, to obtain $B(x, t)$, we use a deep neural network (DNN) to represent the solution of~(\ref{pde-cbvf}).
The network $B_{\theta}(x, t)$ is parameterized by $\theta$ and takes the state $x$ and time $t$ as inputs, and tries to output the value of $B(x, t)$.
The advantage of using a DNN is to avoid the spatial discretization of the state and thus can generalize to large-scale systems.
In this work, we adopt our DNN training from \emph{DeepReach}~\cite{bansal2021deepreach}.
We first sample an input dataset $\{(x_i, t_i)\}_{i\in \{1, 2, \cdots, M\}}$, where $M$ is the sample size, and consider the loss function $\loss(x_i, t_i; \param)$ in \eqref{eq:loss_function} for a data point $(x_i, t_i)$,
where $\loss_1$ requires that the training data should adhere to the HJ PDE dynamic, $\loss_2$ requires that the value of $B_{\theta}(x, t)$ at initial time $t=0$ should be as close to the ground truth as possible, and $\lambda$ is the tradeoff weight.
Thus, the loss function $h$ encourages the trained network to satisfy the boundary condition ($h_1$) and PDE dynamic ($h_2$).
\vspace{-2mm}
\begin{align}
\label{eq:loss_function}
    \loss(x_i, t_i; \param) & = \loss_1(x_i, t_i; \param) + \lambda \loss_2(x_i, t_i; \param), \loss_1(x_i, t_i; \param) = \|B_{\theta}(x_i, t_i) - \ell_{new}(x_i)\| \mathds{1}(t_i = 0), \nonumber\\
    \loss_2(x_i, t_i; \param) & = \|\min\Big\{-\frac{\partial B_{\theta}(x_i, t_i)}{\partial t} + \textsf{Ham}(x_i, t_i), \ell_{new}(x_i)- B_{\theta}(x_i, t_i)\Big\}\|.
\end{align}
\vspace{-1mm}
We conduct the training in three stages.
First, we train the network only using the loss term $h_1$, i.e., the network tries to fit the boundary condition of PDE at $t=0$ at the beginning phase. 
Then, we add loss term $h_2$ back and fit  function $B_\theta$ gradually from $t=0$ to $t=T$, where $T$ is terminal time of neural network CBVF. 
Finally, we decrease the learning rate and train the network for the whole time interval $[0,T]$. 
The terminal time $T$ of CBVF should be infinite theoretically, but it suffices to let $T$ be a finite time in practice, e.g., the pre-set maximum experiment time.
\begin{wrapfigure}{r}{0.61\textwidth}
\begin{minipage}
{0.61\textwidth}
\vspace{-6mm}
    \begin{algorithm}[H]
    \caption{Learning to Refine CBF}
    \label{alg:train}
    
    1. Initialize the CBVF network $B_{\theta}(x, t)$ with parameter $\theta$\;

    2. Sample a training dataset $\mathcal{D}=\{(x_i, t_i)\}_{i\in \{1, 2, \cdots, M\}}$\;

    3. Train $B_{\theta}(x, t)$ using the loss term $h_1$ to fit boundary\;

    4. Train $B_{\theta}(x, t)$ using loss $h$ gradually from $t=0$ to $t=T$ \;
    
    5. Decrease the learning rate, train $B_{\theta}(x, t)$ for all $t\in [0, T]$\;

\end{algorithm}
  \end{minipage}
\end{wrapfigure} 
Note that a straightforward approach to train $B_{\theta}(x, t)$ is supervised learning,
i.e., we collect the ground truth values $y_i = B(x_i, t_i)$ for all sampled data inputs.
However, these ground truths are not available especially for high-dimensional systems.
Thus we resort to self-supervision training method using the loss function~(\ref{eq:loss_function}) then the ground truths of dataset are not necessary.

One challenge for training $B_{\theta}(x, t)$ is that the network should recover not only the values of CBVF, but also the gradient values of CBVF because the gradient is required to compute the safe control inputs during inference.
Leveraging ideas from~\cite{sitzmann2020implicit, bansal2021deepreach}, we use periodic activation functions for our DNN (e.g., sinusoidal function), which are shown to be effective for representing target complex signals and their derivatives.
In addition, to improve the training efficiency, we initialize the safety boundary using the information from initial local CBF $h(x)$ rather than  the constraint function $\ell(x)$, i.e., $\ell_{new}(x_i)$ is replaced by $\min(\ell_{\unsafesw}(x_i), h(x_i))$ in the loss function~(\ref{eq:loss_function}).
This can be considered as a warmstarting since $h(x)$ is typically a better guess of our desired CBVF than $\ell(x)$.
Finally, we can ensure transition safety in Problem~\ref{problem} by controlling the system with $B_{\theta}(x, t)$ at mode $q$ and with $h_{q'}$ at mode $q'$.



\vspace{-4mm}
\subsection{Application in Multi-Modes Systems}
\vspace{-2mm}
\begin{figure}[h!]
  \centering
  \begin{minipage}[b]{0.2\textwidth}
    \centering
    \includegraphics[width=33mm]{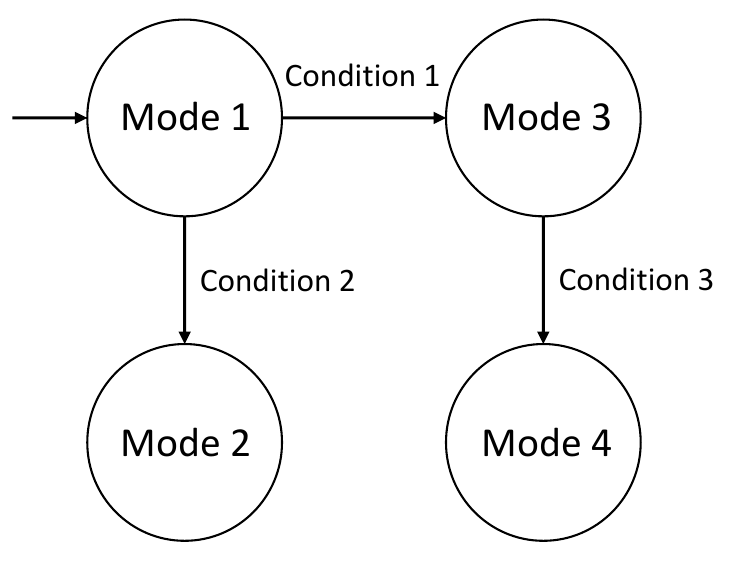}\\
    (a)
  \end{minipage}
  \begin{minipage}[b]{0.2\textwidth}
    \hspace*{0.06cm}
    \centering
     \includegraphics[width=30mm]{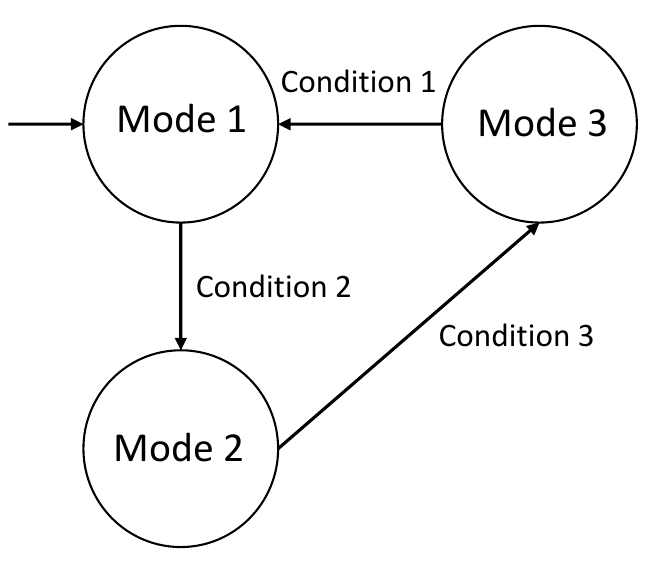}\\
    (b)
  \end{minipage}
  \begin{minipage}[b]{0.2\textwidth}
    \hspace*{0.06cm}
    \centering
    \includegraphics[width=34mm]{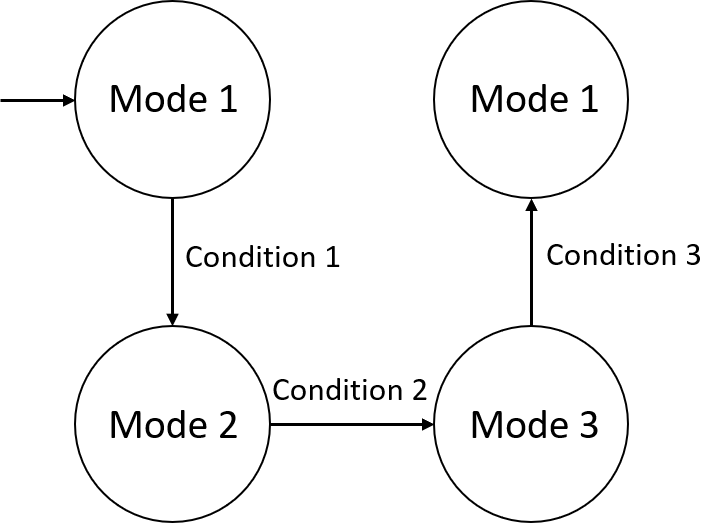}\\
    (c)
  \end{minipage}
  \begin{minipage}[b]{0.2\textwidth}
    \hspace*{0.11cm}
    \centering
    \includegraphics[width=33mm]{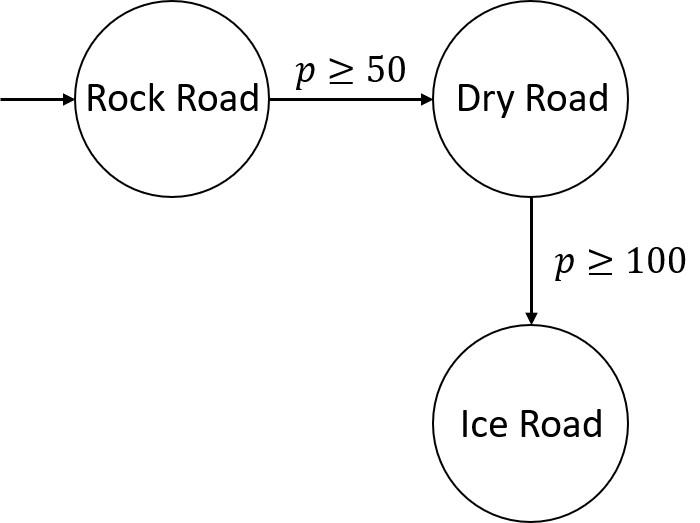}\\
    (d)
  \end{minipage}
   \caption{\small{\textbf{(a)} \scriptsize{Hybrid System 1: no transition cycle.} \textbf{(b)} \scriptsize{Hybrid System 2: with a transition cycle.} \textbf{(c)} \scriptsize{An unfolded system from System 1.} 
   \textbf{(d)} Hybrid adaptive cruise control system. \revise{The switching conditions are when the position $p\ge 50$ and $p\ge 100$.}}}
   \label{fig:system_cycle_examples-and-case-1-system}
\end{figure}
\revise{
We presented how to guarantee transition safety above, i.e., one can learn a CBVF by considering the unsafe switching set for transition between two modes.
Now we discuss how to ensure global safety with multiple transition pairs $\mathcal{T}(H)$.  Note that $\mathcal{T}(H)$ induces a directed graph such that the vertex set of the graph is the discrete operation modes set $Q$ of hybrid system and an edge $(q,q')$ is defined in the graph if $(q,q') \in \mathcal{T}(H)$. 
To analyze how to apply the method we introduced above to hybrid systems with multiple transition paris, we divide the systems into two cases:
}

\noindent\textbf{Case 1.}
If the graph is \emph{acyclic}, i.e., it is a tree, e.g., see Figure~(\ref{fig:system_cycle_examples-and-case-1-system}-a). 
We can ensure global safety by enforcing transition safety from leaf nodes to root nodes in this case. 
Regarding the refinement complexity, since we only need to refine at most one time for each mode and leaf modes do not have to be refined, so we only have to refine $(\texttt{Num}_m-\texttt{Num}_l)$ times, where $\texttt{Num}_m$ is the mode number and $\texttt{Num}_l$ is the leaf mode number.
Thus, the complexity is linear w.r.t. the mode number.

\noindent \textbf{Case 2.}
If the graph is \emph{cyclic}, i.e., it contains some loops, e.g., see Figure~(\ref{fig:system_cycle_examples-and-case-1-system}-b). 
In general, we need to iteratively compute unsafe backward sets to refine the local CBF for each mode
switching until their convergence. 
This refinement may take infinite time in the worst case.
Proposing an efficient algorithm with convergence guarantees in this case is rather challenging and we leave it as a future direction.
However, if the system task is defined in terms of modes, e.g., if the task in Figure~(\ref{fig:system_cycle_examples-and-case-1-system}-b) is defined as executing Mode 1, Mode 2, Mode 3 sequentially, and finally back to Mode 1, then we can unfold the graph until the task is finished. 
The unfolded graph is shown in Figure~(\ref{fig:system_cycle_examples-and-case-1-system}-c), which belongs to Case 1 now since it is a tree.
By refining the unfolded tree, we can still ensure safety for the specific task.
We will demonstrate these two cases through two experiments respectively.




\vspace{-5mm}
\section{Numerical Experiments}\label{sec:experiment}
\vspace{-2mm}
In this section, we illustrate the proposed approach by numerical experiments. 
First, we use the adaptive cruise control (ACC) problem to demonstrate the efficiency of our approach compared with existing approaches.
Then, we showcase simulations on a high-dimensional 2D autonomous racing case, which cannot be addressed by HJ-reach-based approach proposed in~\cite{yang2023safe}. 
In particular, \revise{we} consider and compare the following control methods:
\begin{figure}[h!]
  \centering
  \begin{minipage}[b]{0.3\textwidth}
    \centering
    \includegraphics[width=47mm]{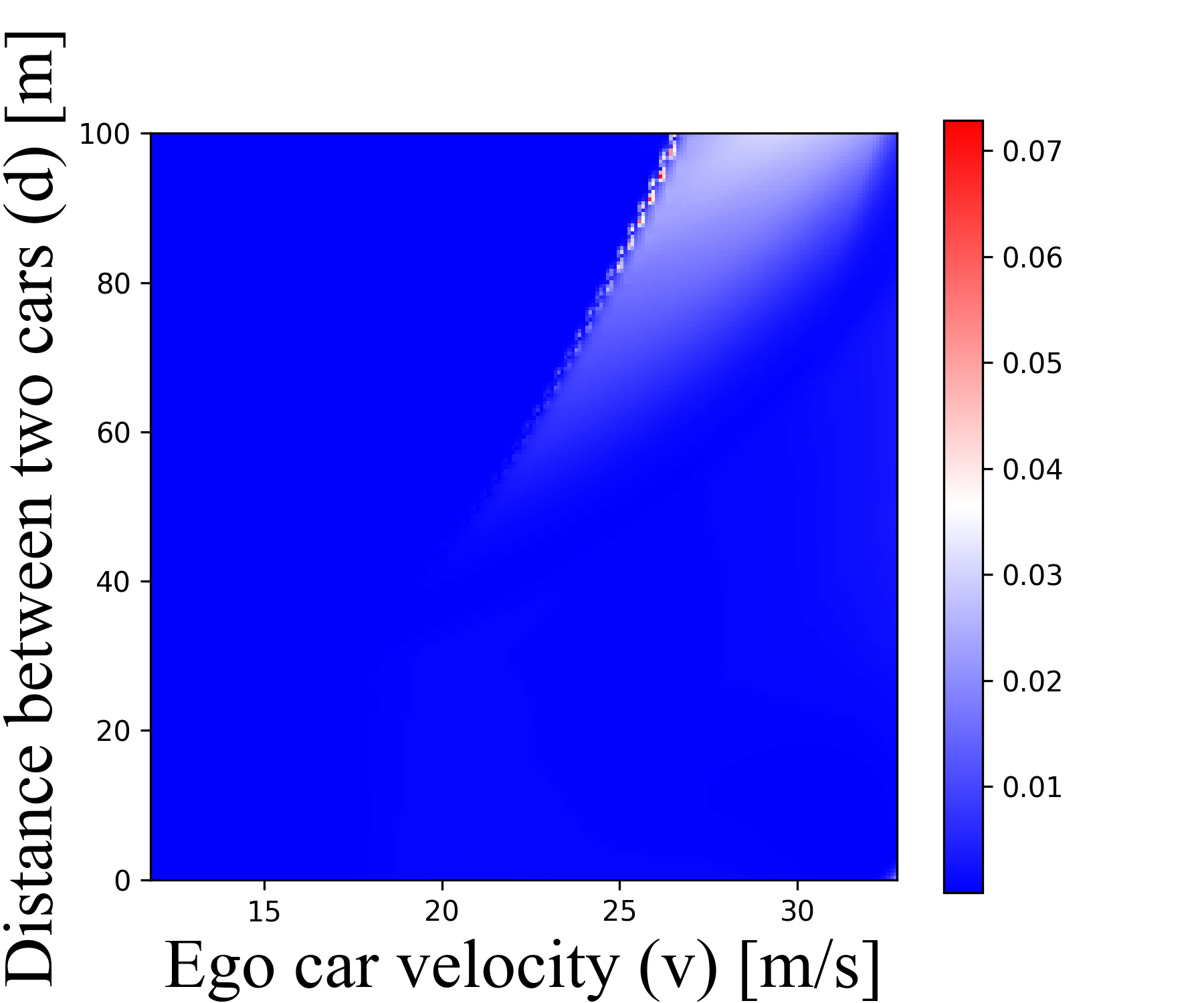}\\
    (a)
  \end{minipage}
  \begin{minipage}[b]{0.3\textwidth}
    \hspace*{0.06cm}
    \centering
     \includegraphics[width=45mm]{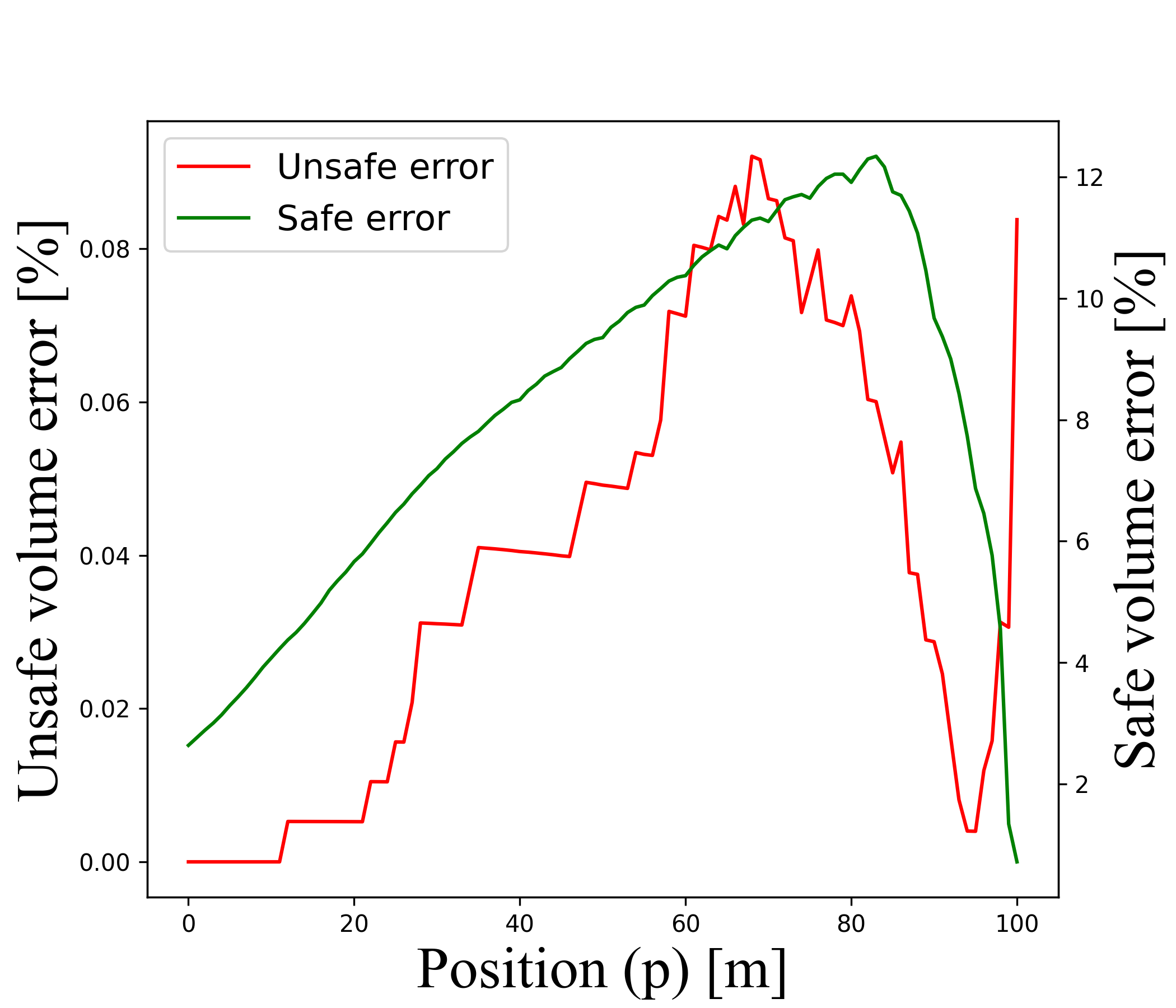}\\
    (b)
  \end{minipage}
  \begin{minipage}[b]{0.3\textwidth}
    \hspace*{0.06cm}
    \centering
    \includegraphics[width=51mm]{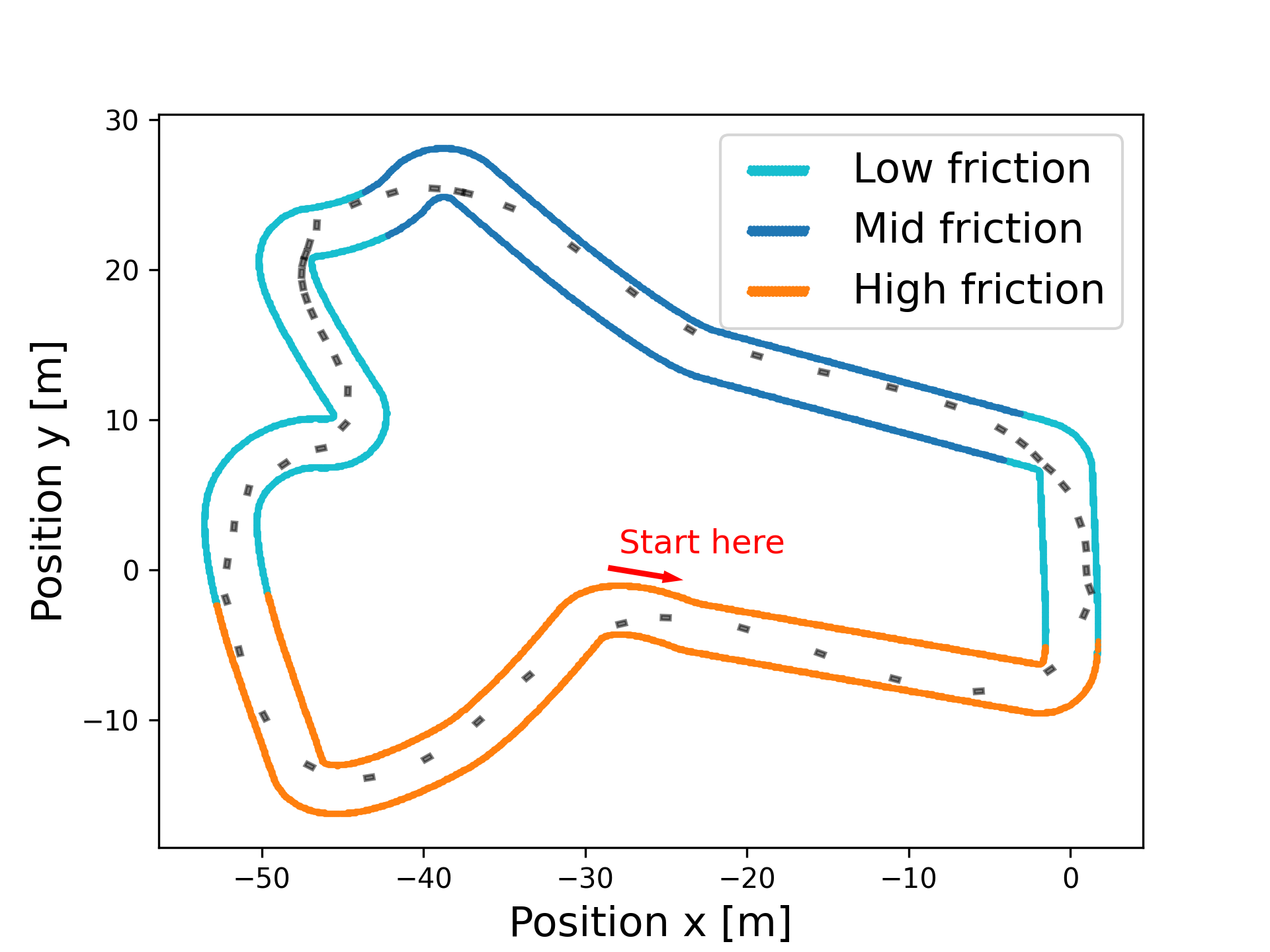}\\
    (c)
  \end{minipage}
   \caption{\small{\textbf{(a)} \scriptsize{Average mean square error of training outcome.} \textbf{(b)} \scriptsize{Safe and unsafe volume errors.} \textbf{(c)} \scriptsize{Multi-friction racing track. The trajectory from our approach is also presented.}}}
   \label{fig:volumn_error_and_racing_track}
\end{figure}
\vspace{-1mm}
\begin{itemize}[itemsep=2pt,topsep=0pt,parsep=0pt]
      \item \textbf{Neural Switch-Aware CBFs (our method)}: We use refined neural CBFs to be aware of safe switching and also be applicable to large-scale systems.
    \item \textbf{Global CBF}~\cite{lindemann2021learning}: Global CBF can also guarantee  safety for  hybrid systems. However, it is inherently difficult to construct such a global CBF and also poses more restrictions than local CBFs-based method, i.e., global CBF deteriorates system performance.
    \item \textbf{Switch-Aware CBFs} from~\cite{yang2023safe}: it can guarantee safety but is limited to small-scale systems.  In our ACC scenario, we consider this approach as an ``optimal'' CBF-based approach and show that performance of our method is close to it.
    \item \textbf{Switch-Unaware CBFs}: 
    Each initial local CBF is applied directly at each mode. This approach is unaware of the safety after switching, and therefore,  safety cannot be guaranteed.
    \item \textbf{MPC}: 
    Model predictive control can guarantee safety of the closed-loop system, but is computationally expensive as it involves solving a receding-horizon optimization online. When the dynamics \revise{are} nonlinear or the horizon is large, the computation cost is even higher.
\end{itemize}
The metrics compared on are \emph{system performance} and \emph{online solving time}, where system performance is related to the case-specific objective.
Our simulations are implemented in \texttt{Python 3} with \texttt{CasADi}~\cite{andersson2019casadi} as the nonlinear optimization problems solver.
Videos can be accessed online at \url{https://youtu.be/aHg0p6zyGFg}.
Additional experimental details and results can be found in the Appendix.
\vspace{-5mm}
\subsection{Adaptive Cruise Control}
\vspace{-2mm}
\noindent\textbf{Scenario Setup.} Adaptive cruise control (ACC) is a common example to validate safe control strategies~\cite{ames2016control, zeng2021safety, yang2022differentiable}, where the ego car and the leading car are driving on a straight road and the ego car is expected to maintain a safe distance with the leading car.
In this case study, we consider ACC with 
\revise{multi-frictions road, which consists of rock road, dry road and ice road, as shown in Figure~(\ref{fig:system_cycle_examples-and-case-1-system}-d).}
We use the same road model as~\cite{yang2023safe} with different friction coefficients for each road mode.
In our simulation, we have the safety specification defined by the constraint function $c(x)=d-T_hv$, where $d$ is the distance between two cars, $v$ is the ego car speed, and $T_h$ is the look-ahead time.
The leading car is driving with speed $v_0$ and the desired cruise speed for ego car is $v_d>v_0$.
We use local CBFs  from~\cite{tonkens2022refining} with different friction coefficients for each road mode.
The control objective is that the ego car is expected to achieve the desired speed $v_{d}$ and to follow the leading car as close as possible.
Specifically, the trajectory cost is defined as:
$\texttt{Cost}=\int_{0}^T (0.01(v(t) - v_{d})^2 + 0.1(d(t) - T_hv(t))^2) dt.$

\noindent\textbf{Training Details.} 
\revise{The mode graph of ACC is acyclic and we train the CBVF network for dry road first, and then train the CBVF network for rock road considering the new safe set of dry road dynamic.
The trainings for two networks are the same following Algorithm~(\ref{alg:train}).}
Overall training time is around $1.5$ hour  using PyTorch with an NVIDIA GeForce RTX3090. 
The training dataset consists of 65K sampled states,  
among which 25K states are close to the switching regions between roads in order to increase the approximation accuracy over the switching area.

\noindent\textbf{Training Results.}
Our trained DNN can represent the PDE solution very accurately compared with the ground truth from~\cite{yang2023safe}.
The training results \revise{of dry road CBVF training} are presented in Figure~(\ref{fig:volumn_error_and_racing_track}). 
We compute the mean square error (MSE) over the position of ego car in Figure~(\ref{fig:volumn_error_and_racing_track}-a).
Also, the calculated MSE over all dimensions is $1.9 \times 10^{-3}$. 
Additionally, we show the unsafe volume error (UVE) and safe volume error (SVE) in Figure~(\ref{fig:volumn_error_and_racing_track}-b), in which UVE is the percentage volume of unsafe states that are considered safe by DNN and SVE is the percentage volume of safe states that are considered unsafe by DNN. 
We can observe that UVE is less than $0.1\%$ and SVE is less than $12\%$, which implies that our trained DNN is very safe but also a bit conservative.
We numerically show that the performance degradation is actually slight soon.


\noindent\textbf{Numerical Comparison.} We test four different baseline approaches over a horizon $T_{sim} = 20s$ to evaluate their system performances. For CBF-based approaches, we choose a PID controller as the reference controller. 
The results are reported in Table~\ref{table:acc_atuto_results}.
We can observe that our neural switch-aware CBFs achieve very close performance to vanilla switch-aware CBFs~\cite{yang2023safe}, and it is even comparable with MPC. 
Switch-unaware local CBFs violates hybrid system safety.
Global CBF method is more conservative than local CBFs-based methods.
In terms of online evaluation time, CBF-based approaches enjoy clear advantages over MPC. 

\begin{threeparttable}
\centering
\small
\caption{\small{Comparisons of different methods. 
The performance metrics of ACC and racing are trajectory cost and lap time (s) respectively.}}
\vspace{-3mm}
\label{table:acc_atuto_results}
\begin{tabular}{ccccccc}
\toprule
  & \multicolumn{2}{c}{Safety} & \multicolumn{2}{c}{Performance} & \multicolumn{2}{c}{Solve time per step (s)} \\
      \cmidrule(lr){2-3}\cmidrule(lr){4-5}\cmidrule(lr){6-7}
      & ACC & Racing & ACC  & Racing & ACC & Racing \\ 

\midrule
MPC & Safe & Safe & 201.2 & 34.3 & $1.7\!\times\! 10^{-1}$ & $7.2\!\times\! 10^{-1}$ \\
Global CBF & Safe & Safe & 329.7 & 69.2 & $5.6\!\times\! 10^{-3}$ & $8.4\!\times\! 10^{-3}$ \\
Switch-unaware CBFs & Unsafe & Unsafe & 250.7 & N/A & $5.7\!\times\! 10^{-3}$ & $8.2\!\times\! 10^{-3}$\\
Switch-aware CBFs~\cite{yang2023safe}  & Safe & N/A & 228.6 & N/A & $5.7\!\times\! 10^{-3}$ & N/A\\
\textbf{Neural switch-aware CBFs}  & Safe & Safe & 236.2 & 42.4 & $5.7\!\times\! 10^{-3}$ & $8.0\!\times\! 10^{-3}$\\
\bottomrule
\end{tabular}
\vspace{-3mm}
\end{threeparttable}

\subsection{2D Autonomous Racing}
\noindent\textbf{Scenario Setup.}
We consider a second case study using the \texttt{F1Tenth}~\cite{o2020f1tenth} with single-track model~\cite{althoff2017commonroad}.
The state space model of single-track model is of 7 dimensions with two control inputs.
Note that HJ-reach practically can only handle system with state dimension less than 6, so this case study cannot be addressed by approach in~\cite{yang2023safe}.
The racing track is a multi-friction road as shown in Figure~(\ref{fig:volumn_error_and_racing_track}-c).
In this context, our safety specification is no collision with the wall.
To ensure the safety, we use a CBF from~\cite{berducci2023learning} which has friction-dependent coefficients.
The control objective is to track a pre-computed optimal racing line~\cite{heilmeier2019minimum} in the map and finish one lap while avoiding collision with the boundaries.
\revise{Note that even if this task is cyclic, we can unfold the cyclic mode transitions into a tree, similar to the example in Figure~(\ref{fig:system_cycle_examples-and-case-1-system}-c), then the safety refinement is tractable and can be guaranteed.}
For CBF-based approaches, we use pure pursuit and PID as our reference planner and controller.

\noindent\textbf{Results.} We train two DNNs (for the left and right walls respectively) with three stages similar to the case of the ACC example. 
The simulation results are reported in Table~\ref{table:acc_atuto_results}.
Vanilla switch-aware CBFs~\cite{yang2023safe} is not applicable due to the high dimensionality of system state.
MPC can finish one lap fastest while preserving safety, but its solving time per step is as high as almost 1 second.
Our neural switch-aware CBFs performs close to MPC and the solving time is significantly lower.
Global CBF is conservative regarding lap time and switch-unaware CBFs is not safe.


\vspace{-3mm}
\section{Conclusion and Discussion}
In this work, we proposed a learning-enabled local CBFs method for safety control of hybrid systems.
Specifically, we leveraged learning techniques to refine existing local CBFs, which are constructed without considering mode switchings, to ensure global safety.  Our approach can scale to high-dimension systems and enjoys low computation cost during inference time. 
The proposed approach outperforms other safety-critical control methods, which is illustrated by two numerical simulations including one high-dimension autonomous racing case. In  the future, we are interested in providing theoretical guarantees for the trained DNN to assure the validity of our neural CBFs. 

\bibliography{reference}

\begin{thebibliography}{41}
\providecommand{\natexlab}[1]{#1}
\providecommand{\url}[1]{\texttt{#1}}
\expandafter\ifx\csname urlstyle\endcsname\relax
  \providecommand{\doi}[1]{doi: #1}\else
  \providecommand{\doi}{doi: \begingroup \urlstyle{rm}\Url}\fi

\bibitem[Abate et~al.(2021)Abate, Ahmed, Edwards, Giacobbe, and Peruffo]{abate2021fossil}
Alessandro Abate, Daniele Ahmed, Alec Edwards, Mirco Giacobbe, and Andrea Peruffo.
\newblock {FOSSIL}: a software tool for the formal synthesis of lyapunov functions and barrier certificates using neural networks.
\newblock In \emph{Proc. of the 24th Int. conf. on Hyb. Syst.: Comput. and Control}, pages 1--11, 2021.

\bibitem[Althoff et~al.(2017)Althoff, Koschi, and Manzinger]{althoff2017commonroad}
Matthias Althoff, Markus Koschi, and Stefanie Manzinger.
\newblock Common{R}oad: Composable benchmarks for motion planning on roads.
\newblock In \emph{IEEE Intell. Vehicles Symposium}, pages 719--726. IEEE, 2017.

\bibitem[Ames et~al.(2016)Ames, Xu, Grizzle, and Tabuada]{ames2016control}
Aaron~D Ames, Xiangru Xu, Jessy~W Grizzle, and Paulo Tabuada.
\newblock Control barrier function based quadratic programs for safety critical systems.
\newblock \emph{IEEE Trans.\ on Automatic Control}, 62\penalty0 (8):\penalty0 3861--3876, 2016.

\bibitem[Ames et~al.(2019)Ames, Coogan, Egerstedt, Notomista, Sreenath, and Tabuada]{ames2019control}
Aaron~D Ames, Samuel Coogan, Magnus Egerstedt, Gennaro Notomista, Koushil Sreenath, and Paulo Tabuada.
\newblock Control barrier functions: Theory and applications.
\newblock In \emph{18th European Control conf.}, pages 3420--3431. IEEE, 2019.

\bibitem[Andersson et~al.(2019)Andersson, Gillis, Horn, Rawlings, and Diehl]{andersson2019casadi}
Joel~AE Andersson, Joris Gillis, Greg Horn, James~B Rawlings, and Moritz Diehl.
\newblock Casadi: a software framework for nonlinear optimization and optimal control.
\newblock \emph{Math. Programming Comput.}, 11:\penalty0 1--36, 2019.

\bibitem[Bansal and Tomlin(2021)]{bansal2021deepreach}
Somil Bansal and Claire~J Tomlin.
\newblock Deepreach: A deep learning approach to high-dimensional reachability.
\newblock In \emph{IEEE Int. conf. on Robotics and Automation}, pages 1817--1824. IEEE, 2021.

\bibitem[Bansal et~al.(2017)Bansal, Chen, Herbert, and Tomlin]{bansal2017hamilton}
Somil Bansal, Mo~Chen, Sylvia Herbert, and Claire~J Tomlin.
\newblock Hamilton-jacobi reachability: A brief overview and recent advances.
\newblock In \emph{56th IEEE conf. on Decis. and Control}, pages 2242--2253. IEEE, 2017.

\bibitem[Benerecetti et~al.(2013)Benerecetti, Faella, and Minopoli]{benerecetti2013automatic}
Massimo Benerecetti, Marco Faella, and Stefano Minopoli.
\newblock Automatic synthesis of switching controllers for linear hybrid systems: Safety control.
\newblock \emph{Theoretical Computer Science}, 493:\penalty0 116--138, 2013.

\bibitem[Berducci et~al.(2024)Berducci, Yang, Mangharam, and Grosu]{berducci2023learning}
Luigi Berducci, Shuo Yang, Rahul Mangharam, and Radu Grosu.
\newblock Learning adaptive safety for multi-agent systems.
\newblock In \emph{IEEE Int. conf. on Robotics and Automation}. IEEE, 2024.
\newblock arXiv:2309.10657.

\bibitem[Berg and Nystr{\"o}m(2018)]{berg2018unified}
Jens Berg and Kaj Nystr{\"o}m.
\newblock A unified deep artificial neural network approach to partial differential equations in complex geometries.
\newblock \emph{Neurocomputing}, 317:\penalty0 28--41, 2018.

\bibitem[Borrelli et~al.(2017)Borrelli, Bemporad, and Morari]{borrelli2017predictive}
Francesco Borrelli, Alberto Bemporad, and Manfred Morari.
\newblock \emph{Predictive control for linear and hybrid systems}.
\newblock Cambridge University Press, 2017.

\bibitem[Choi et~al.(2021)Choi, Lee, Sreenath, Tomlin, and Herbert]{choi2021robust}
Jason~J Choi, Donggun Lee, Koushil Sreenath, Claire~J Tomlin, and Sylvia~L Herbert.
\newblock Robust control barrier--value functions for safety-critical control.
\newblock In \emph{60th IEEE conf. on Decis. and Control}, pages 6814--6821. IEEE, 2021.

\bibitem[Evans et~al.(2023)Evans, Engelbrecht, and Jordaan]{evans2023high}
Benjamin~David Evans, Herman~Arnold Engelbrecht, and Hendrik~Willem Jordaan.
\newblock High-speed autonomous racing using trajectory-aided deep reinforcement learning.
\newblock \emph{IEEE Robotics and Automation Letters}, 2023.

\bibitem[Glotfelter et~al.(2019)Glotfelter, Buckley, and Egerstedt]{glotfelter2019hybrid}
Paul Glotfelter, Ian Buckley, and Magnus Egerstedt.
\newblock Hybrid nonsmooth barrier functions with applications to provably safe and composable collision avoidance for robotic systems.
\newblock \emph{IEEE Robotics and Automation Letters}, 4\penalty0 (2):\penalty0 1303--1310, 2019.

\bibitem[Goebel et~al.(2009)Goebel, Sanfelice, and Teel]{goebel2009hybrid}
Rafal Goebel, Ricardo~G Sanfelice, and Andrew~R Teel.
\newblock Hybrid dynamical systems.
\newblock \emph{IEEE Control Syst. Mag.}, 29\penalty0 (2):\penalty0 28--93, 2009.

\bibitem[Han et~al.(2018)Han, Jentzen, and E]{han2018solving}
Jiequn Han, Arnulf Jentzen, and Weinan E.
\newblock Solving high-dimensional partial differential equations using deep learning.
\newblock \emph{Proc. of the National Academy of Sciences}, 115\penalty0 (34):\penalty0 8505--8510, 2018.

\bibitem[Heilmeier et~al.(2019)Heilmeier, Wischnewski, Hermansdorfer, Betz, Lienkamp, and Lohmann]{heilmeier2019minimum}
Alexander Heilmeier, Alexander Wischnewski, Leonhard Hermansdorfer, Johannes Betz, Markus Lienkamp, and Boris Lohmann.
\newblock Minimum curvature trajectory planning and control for an autonomous race car.
\newblock \emph{Vehicle System Dynamics}, 2019.

\bibitem[Hsu et~al.(2023)Hsu, Hu, and Fisac]{hsu2023safety}
Kai-Chieh Hsu, Haimin Hu, and Jaime~Fern{\'a}ndez Fisac.
\newblock The safety filter: A unified view of safety-critical control in autonomous systems.
\newblock \emph{arXiv preprint arXiv:2309.05837}, 2023.

\bibitem[Ivanov et~al.(2019)Ivanov, Weimer, Alur, Pappas, and Lee]{ivanov2019verisig}
Radoslav Ivanov, James Weimer, Rajeev Alur, George~J Pappas, and Insup Lee.
\newblock Verisig: verifying safety properties of hybrid systems with neural network controllers.
\newblock In \emph{Proc. of the 22nd ACM Int. conf. on Hyb. Syst.: Comput. and Control}, pages 169--178, 2019.

\bibitem[Legat et~al.(2018)Legat, Tabuada, and Jungers]{legat2018computing}
Beno{\^\i}t Legat, Paulo Tabuada, and Rapha{\"e}l~M Jungers.
\newblock Computing controlled invariant sets for hybrid systems with applications to model-predictive control.
\newblock \emph{IFAC-PapersOnLine}, 51\penalty0 (16):\penalty0 193--198, 2018.

\bibitem[Lindemann et~al.(2021)Lindemann, Hu, Robey, Zhang, Dimarogonas, Tu, and Matni]{lindemann2021learning}
Lars Lindemann, Haimin Hu, Alexander Robey, Hanwen Zhang, Dimos Dimarogonas, Stephen Tu, and Nikolai Matni.
\newblock Learning hybrid control barrier functions from data.
\newblock In \emph{conf. on Robot Learning}, pages 1351--1370. PMLR, 2021.

\bibitem[Maghenem and Sanfelice(2019)]{maghenem2019characterizations}
Mohamed Maghenem and Ricardo~G Sanfelice.
\newblock Characterizations of safety in hybrid inclusions via barrier functions.
\newblock In \emph{Proc. of the 22nd ACM Int. conf. on Hyb. Syst.: Comput. and Control}, pages 109--118, 2019.

\bibitem[Mhaskar et~al.(2005)Mhaskar, El-Farra, and Christofides]{mhaskar2005robust}
Prashant Mhaskar, Nael~H El-Farra, and Panagiotis~D Christofides.
\newblock Robust hybrid predictive control of nonlinear systems.
\newblock \emph{Automatica}, 41\penalty0 (2):\penalty0 209--217, 2005.

\bibitem[Nejati et~al.(2022)Nejati, Soudjani, and Zamani]{nejati2022compositional}
Ameneh Nejati, Sadegh Soudjani, and Majid Zamani.
\newblock Compositional construction of control barrier functions for continuous-time stochastic hybrid systems.
\newblock \emph{Automatica}, 145:\penalty0 110513, 2022.

\bibitem[O'Kelly et~al.(2020)O'Kelly, Zheng, Karthik, and Mangharam]{o2020f1tenth}
Matthew O'Kelly, Hongrui Zheng, Dhruv Karthik, and Rahul Mangharam.
\newblock F1tenth: An open-source evaluation environment for continuous control and reinforcement learning.
\newblock \emph{Proc. of Mac. Learning Res.}, 123, 2020.

\bibitem[Phan et~al.(2019)Phan, Paoletti, Zhang, Grosu, Smolka, and Stoller]{phan2019neural}
Dung Phan, Nicola Paoletti, Timothy Zhang, Radu Grosu, Scott~A Smolka, and Scott~D Stoller.
\newblock Neural state classification for hybrid systems.
\newblock In \emph{Proc. of the Fifth Int. Workshop on Symbolic-Numeric Methods for Reasoning About CPS and IoT}, pages 24--27, 2019.

\bibitem[Prajna and Jadbabaie(2004)]{prajna2004safety}
Stephen Prajna and Ali Jadbabaie.
\newblock Safety verification of hybrid systems using barrier certificates.
\newblock In \emph{Int. Workshop on Hyb. Syst.: Comput. and Control}, pages 477--492. Springer, 2004.

\bibitem[Qin et~al.(2021)Qin, Zhang, Chen, Chen, and Fan]{qin2021learning}
Zengyi Qin, Kaiqing Zhang, Yuxiao Chen, Jingkai Chen, and Chuchu Fan.
\newblock Learning safe multi-agent control with decentralized neural barrier certificates.
\newblock In \emph{Int. conf. on Learning Representations}, 2021.
\newblock arXiv:2101.05436.

\bibitem[Robey et~al.(2020)Robey, Hu, Lindemann, Zhang, Dimarogonas, Tu, and Matni]{robey2020learning}
Alexander Robey, Haimin Hu, Lars Lindemann, Hanwen Zhang, Dimos~V Dimarogonas, Stephen Tu, and Nikolai Matni.
\newblock Learning control barrier functions from expert demonstrations.
\newblock In \emph{59th IEEE conf. on Decis. and Control}, pages 3717--3724. IEEE, 2020.

\bibitem[Robey et~al.(2021)Robey, Lindemann, Tu, and Matni]{robey2021learning}
Alexander Robey, Lars Lindemann, Stephen Tu, and Nikolai Matni.
\newblock Learning robust hybrid control barrier functions for uncertain systems.
\newblock \emph{IFAC-PapersOnLine}, 54\penalty0 (5):\penalty0 1--6, 2021.

\bibitem[Sirignano and Spiliopoulos(2018)]{sirignano2018dgm}
Justin Sirignano and Konstantinos Spiliopoulos.
\newblock {DGM}: A deep learning algorithm for solving partial differential equations.
\newblock \emph{Journal of Computational Physics}, 375:\penalty0 1339--1364, 2018.

\bibitem[Sitzmann et~al.(2020)Sitzmann, Martel, Bergman, Lindell, and Wetzstein]{sitzmann2020implicit}
Vincent Sitzmann, Julien Martel, Alexander Bergman, David Lindell, and Gordon Wetzstein.
\newblock Implicit neural representations with periodic activation functions.
\newblock \emph{Adv. in Neural Inf. Processing Syst.}, 33:\penalty0 7462--7473, 2020.

\bibitem[Srinivasan et~al.(2020)Srinivasan, Dabholkar, Coogan, and Vela]{srinivasan2020synthesis}
Mohit Srinivasan, Amogh Dabholkar, Samuel Coogan, and Patricio~A Vela.
\newblock Synthesis of control barrier functions using a supervised machine learning approach.
\newblock In \emph{IEEE Int. conf. on Intell. Robots and Syst.}, pages 7139--7145. IEEE, 2020.

\bibitem[Sun et~al.(2023{\natexlab{a}})Sun, Yang, Zhou, Liu, and Mangharam]{sun2023mega}
Xiatao Sun, Shuo Yang, Mingyan Zhou, Kunpeng Liu, and Rahul Mangharam.
\newblock Mega-dagger: Imitation learning with multiple imperfect experts.
\newblock \emph{arXiv preprint arXiv:2303.00638}, 2023{\natexlab{a}}.

\bibitem[Sun et~al.(2023{\natexlab{b}})Sun, Zhou, Zhuang, Yang, Betz, and Mangharam]{sun2023benchmark}
Xiatao Sun, Mingyan Zhou, Zhijun Zhuang, Shuo Yang, Johannes Betz, and Rahul Mangharam.
\newblock A benchmark comparison of imitation learning-based control policies for autonomous racing.
\newblock In \emph{2023 IEEE Intelligent Vehicles Symposium (IV)}, pages 1--5. IEEE, 2023{\natexlab{b}}.

\bibitem[Tomlin et~al.(2003)Tomlin, Mitchell, Bayen, and Oishi]{tomlin2003computational}
Claire~J Tomlin, Ian Mitchell, Alexandre~M Bayen, and Meeko Oishi.
\newblock Computational techniques for the verification of hybrid systems.
\newblock \emph{Proc. of the IEEE}, 91\penalty0 (7):\penalty0 986--1001, 2003.

\bibitem[Tonkens and Herbert(2022)]{tonkens2022refining}
Sander Tonkens and Sylvia Herbert.
\newblock Refining control barrier functions through hamilton-jacobi reachability.
\newblock In \emph{IEEE Int. conf. on Intell. Robots and Syst.}, pages 13355--13362. IEEE, 2022.

\bibitem[Wang et~al.(2023)Wang, Zhan, Jiao, Wang, Jin, Yang, Wang, Huang, and Zhu]{wang2023enforcing}
Yixuan Wang, Simon~Sinong Zhan, Ruochen Jiao, Zhilu Wang, Wanxin Jin, Zhuoran Yang, Zhaoran Wang, Chao Huang, and Qi~Zhu.
\newblock Enforcing hard constraints with soft barriers: Safe reinforcement learning in unknown stochastic environments.
\newblock In \emph{Int. conf. on Mac. Learning}, pages 36593--36604. PMLR, 2023.

\bibitem[Yang et~al.(2022)Yang, Chen, Preciado, and Mangharam]{yang2022differentiable}
Shuo Yang, Shaoru Chen, Victor~M Preciado, and Rahul Mangharam.
\newblock Differentiable safe controller design through control barrier functions.
\newblock \emph{IEEE Control Syst. Letters}, 7:\penalty0 1207--1212, 2022.

\bibitem[Yang et~al.(2024)Yang, Black, Fainekos, Hoxha, Okamoto, and Mangharam]{yang2023safe}
Shuo Yang, Mitchell Black, Georgios Fainekos, Bardh Hoxha, Hideki Okamoto, and Rahul Mangharam.
\newblock Safe control synthesis for hybrid systems through local control barrier functions.
\newblock In \emph{2024 American Control Conference (ACC)}, pages 344--351. IEEE, 2024.

\bibitem[Zeng et~al.(2021)Zeng, Zhang, Li, and Sreenath]{zeng2021safety}
Jun Zeng, Bike Zhang, Zhongyu Li, and Koushil Sreenath.
\newblock Safety-critical control using optimal-decay control barrier function with guaranteed point-wise feasibility.
\newblock In \emph{Amer. Control conf.}, pages 3856--3863. IEEE, 2021.

\end{thebibliography}
\section{Appendix}
\subsection{Proof of Theorem~\ref{thm:no-backunsafe} and Proposition~\ref{prop:same-cbvf}}

\begin{mythm}\label{thm:no-backunsafe}
    The viability kernel $\revise{\mathcal{S}_{\infty, b}}$ of $\mathcal{L}\setminus\texttt{BackUnsafe}$ equals to viability kernel $\revise{\mathcal{S}_{\infty, u}}$ of $\mathcal{L}\setminus\unsafesw$.
\end{mythm}

\begin{proof}
  First we note that $\mathcal{L}\setminus\texttt{BackUnsafe}\subseteq \mathcal{L}\setminus\unsafesw$, so we have $\mathcal{S}_{\infty, b}\subseteq \mathcal{S}_{\infty, u}$.
    Next, we want to prove $\mathcal{S}_{\infty, u}\subseteq \mathcal{S}_{\infty, b}$, i.e., for any state $x\in \mathcal{S}_{\infty, u}$, $x\in \mathcal{S}_{\infty, b}$ also holds.
    We prove this by contradiction and suppose there exists $x$ such that $x\in \mathcal{S}_{\infty, u}$ but $x\not\in \mathcal{S}_{\infty, b}$.
    Since $x\not\in \mathcal{S}_{\infty, b}$ implies that there exists $T\in [0, \infty)$ and $u(\cdot)\in U_{[0, T]}$ such that $\ell(\xi_{x, 0}^{F_q, u}(T))\not\in \mathcal{L}\setminus\texttt{BackUnsafe}$, i.e., $\ell(\xi_{x, 0}^{F_q, u}(T))\in \texttt{BackUnsafe}$ or $\ell(\xi_{x, 0}^{F_q, u}(T))\not\in \mathcal{L}$.
    On the other hand, $x\in \mathcal{S}_{\infty, u}$ means that $\ell(\xi_{x, 0}^{F_q, u}(T))\in \mathcal{S}_{\infty, u}\subseteq\mathcal{L}\setminus\unsafesw\subseteq \mathcal{L}$, so we conclude $\ell(\xi_{x, 0}^{F_q, u}(T))\in \texttt{BackUnsafe}$.
    This implies that there exists $T'\in (T, \infty)$ and $u'(\cdot)\in U_{[T, T']}$ such that $\ell(\xi_{x, 0}^{F_q, uu'}(T'))\in \unsafesw$ based on the definition of $\texttt{BackUnsafe}$, i.e., the trajectory starting from $x$ can enter the unsafe switching set $\unsafesw$.
    This, however, has contradicted with the assumption that $x\in \mathcal{S}_{\infty, u}\subseteq \mathcal{L}\setminus\unsafesw$.
    Finally, we can conclude that $\mathcal{S}_{\infty, u}\subseteq \mathcal{S}_{\infty, b}$ also holds.
\end{proof}

\begin{mypro}\label{prop:same-cbvf}
    The CBVF $B^{b}_{\gamma}(x, \infty)$ constructed based on the constraint set $\mathcal{L}\setminus\texttt{BackUnsafe}$ shares the same safe set with the CBVF $B^{u}_{\gamma}(x, \infty)$ constructed based on the constraint set $\mathcal{L}\setminus\unsafesw$, i.e., $\mathcal{C}_{B^b_\gamma}(\infty)=\mathcal{C}_{B^u_\gamma}(\infty)$.
\end{mypro}
\begin{proof}
Since CBVF $B_{\gamma}(x, \infty)$ recovers the largest viability kernel,  we have that the largest viability kernels of $\mathcal{L}\setminus\texttt{BackUnsafe}$ and $\mathcal{L}\setminus\unsafesw$ are the same according to Theorem~\ref{thm:no-backunsafe}. Therefore, we know that $\mathcal{C}_{B^b_\gamma}(\infty)=\mathcal{C}_{B^u_\gamma}(\infty)$.
\end{proof}

\subsection{Experimental Details and Results}\label{sec:append_training_results}
\begin{figure}
    \centering
    \includegraphics[width=0.3\linewidth]{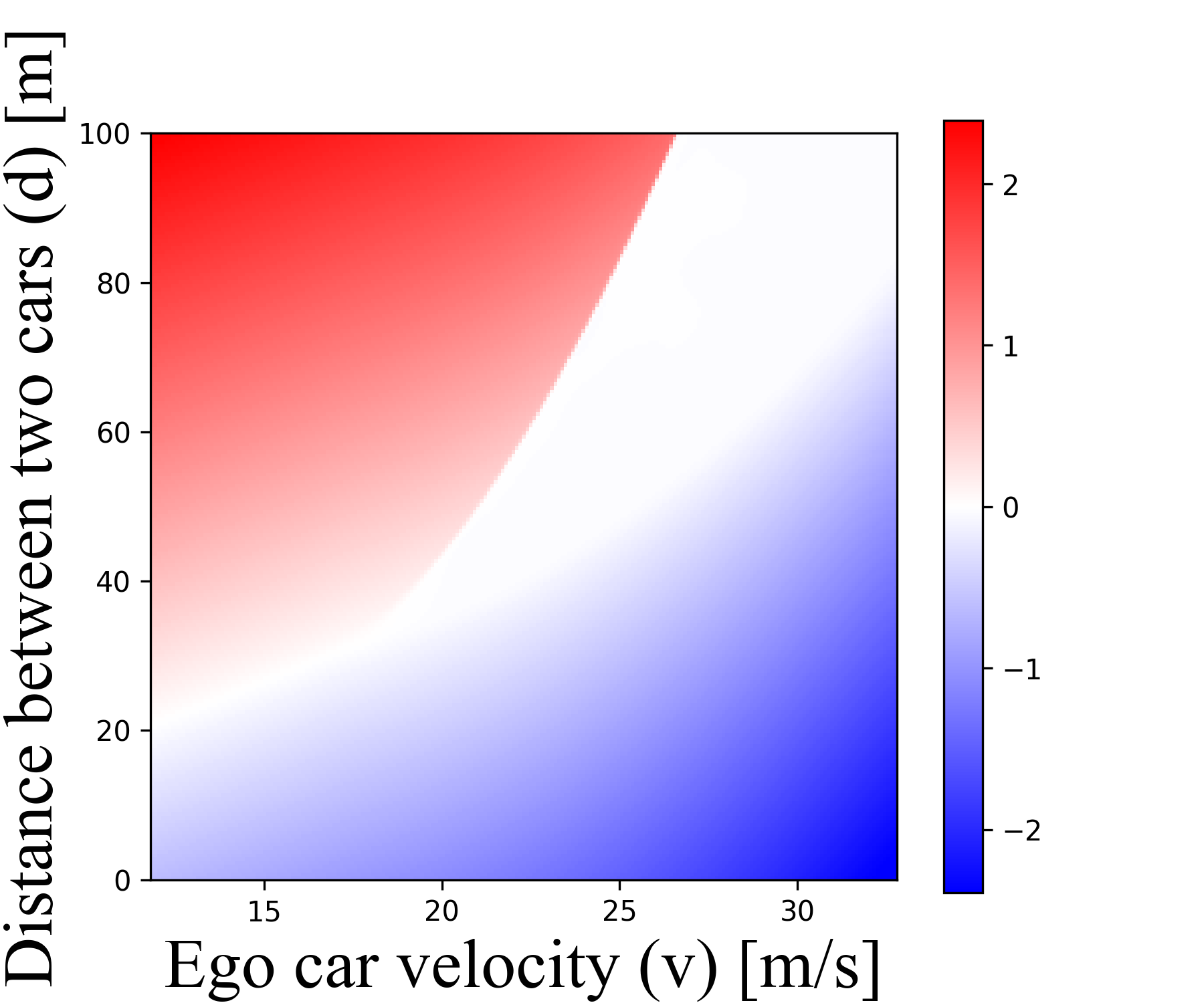}
    \includegraphics[width=0.3\linewidth]{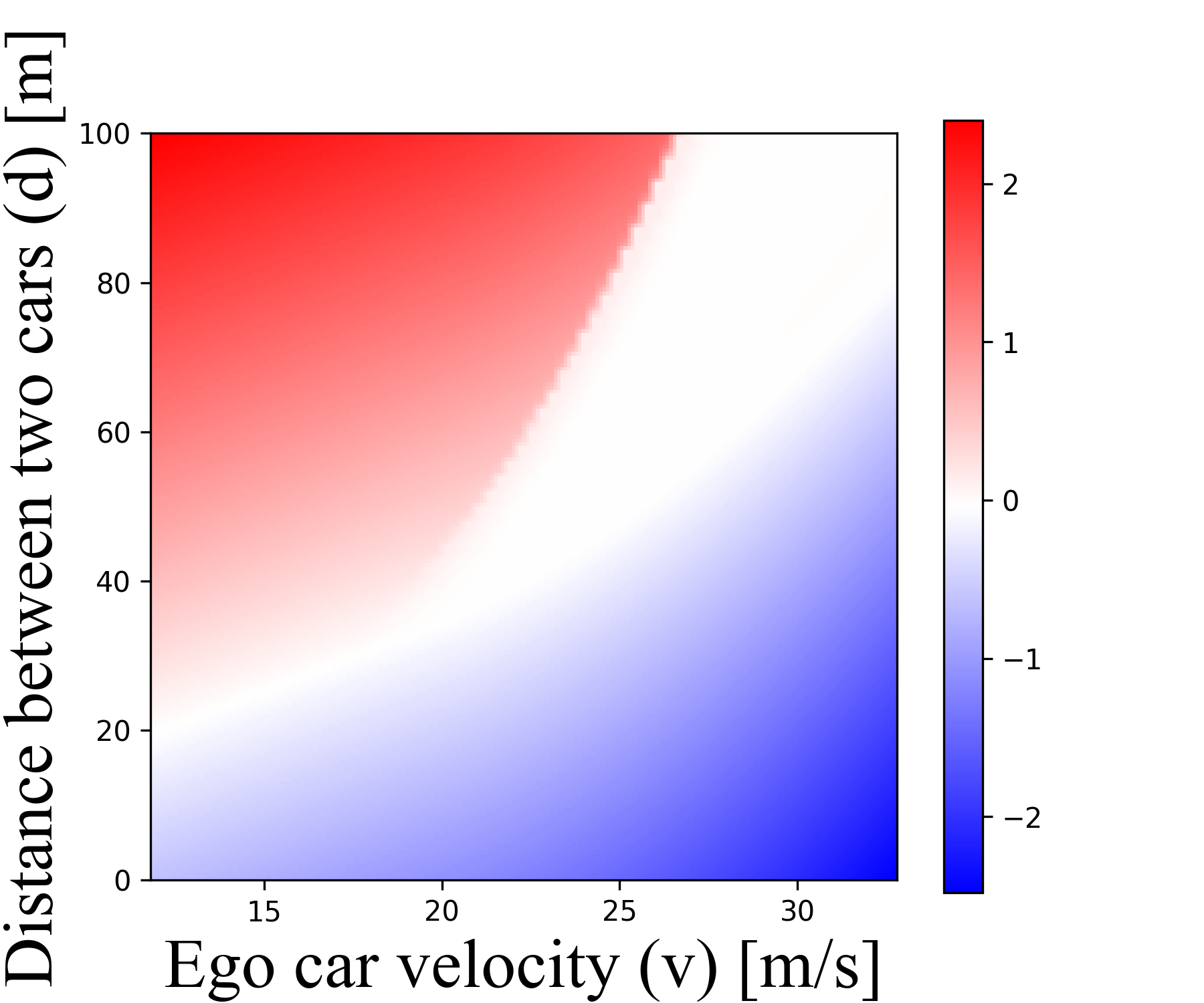}
    \includegraphics[width=0.3\textwidth]{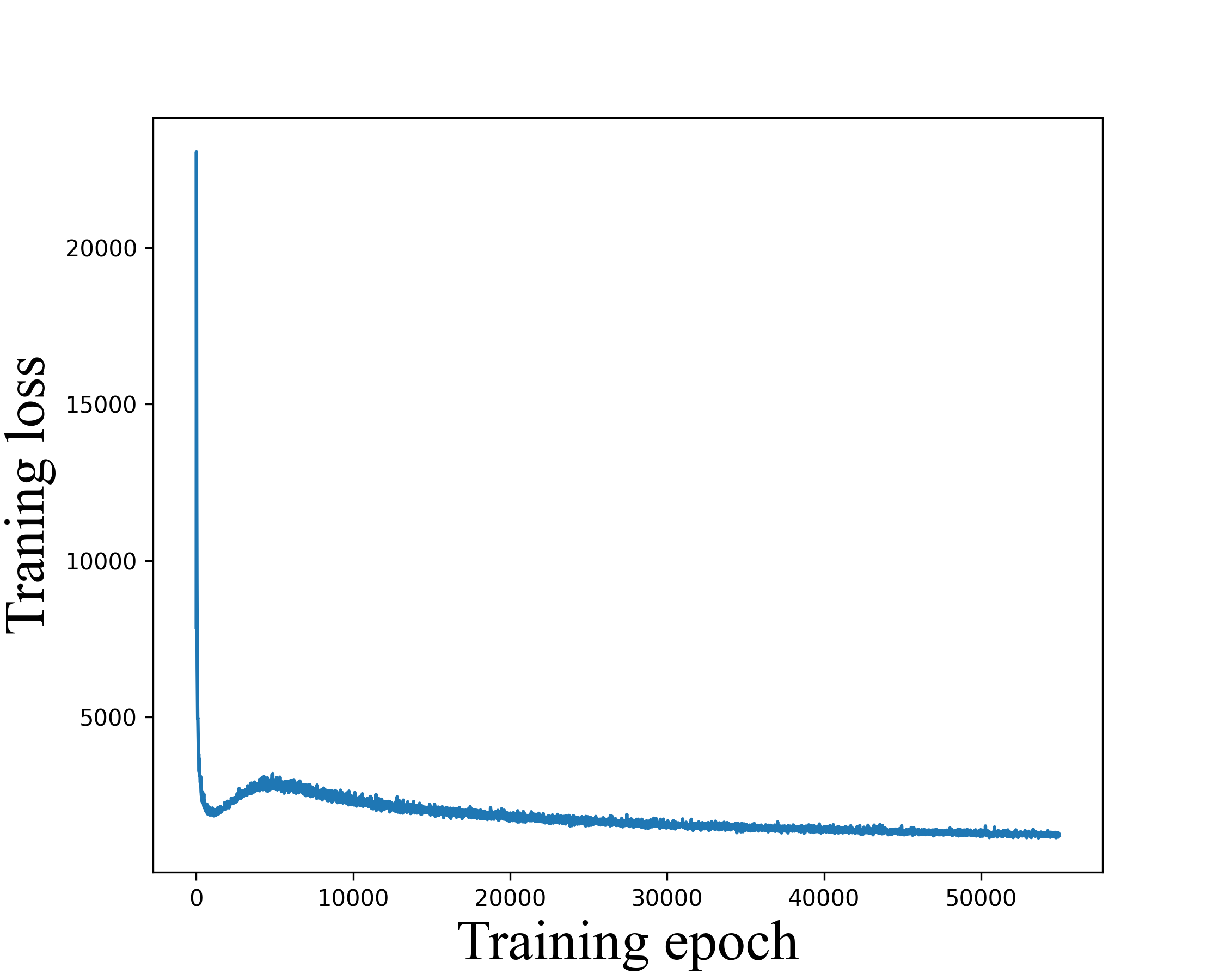}
    
    \caption{Training results of adaptive cruise control. 
    Left figure: DNN value for $p=95$; middle figure: ground truth for $p=95$; right figure: training loss curve.}
    \label{fig:training_outcome}
\end{figure}

\noindent\textbf{Adaptive Cruise Control.}
The training loss curve of dry road CBVF, the DNN and ground truth values when position ($p=95$) is around the switching area ($p=100$), are shown in Figure~\ref{fig:training_outcome}
We can observe that DNN can accurately recover the ground truth value.
Parameters are shown in Table~\ref{table:acc_params}.
Note that the MPC horizon is 70 steps because this is almost the least steps number to ensure feasibility practically.

\noindent\textbf{Autonomous Racing.}
The training procedure is similar to Adaptive Cruise Control case.
Relevant parameters are shown in Table~\ref{table:racing_params}.
Also, 20 steps is almost the minimum horizon for MPC otherwise there is infeasibility issue in practice.
For CBF-based approaches, we use pure pursuit and PID as our reference planner and controller.
One can also use any other potentially unsafe control methods as the reference controllers, for example, imitation learning or reinforcement learning policy~\cite{sun2023mega, sun2023benchmark, evans2023high}.

\subsection{Single-Track Model}\label{appendix: ST-model}
The single-track models a road vehicle with two wheels, where the front and rear wheel pairs are each lumped into one wheel.
Compared with kinematic single-track mode, single-track model additionally considers tire slip, whose effect is more dominant when the vehicle is driving close to the physical capabilities.
The state space model consist of 7 states
$\bold{x} = \begin{bmatrix}
s_x & s_y & \delta & v & \Psi & \dot{\Psi} & \beta
\end{bmatrix}$
where the control input variables are
$
\begin{bmatrix}
u_1 & u_2
\end{bmatrix} = \begin{bmatrix}
v_{\delta} & a_{long}
\end{bmatrix}.
$
\begin{equation}
\begin{aligned}
\dot{x}_1= & x_4 \cos \left(x_5+x_7\right), \\
\dot{x}_2= & x_4 \sin \left(x_5+x_7\right), \\
\dot{x}_3= & f_{steer}\left(x_3, u_1\right), \\
\dot{x}_4= & f_{a c c}\left(x_4, u_2\right) \\
\dot{x}_5= & x_6, \\
\dot{x}_6= & \frac{\mu m}{I_z\left(l_r+l_f\right)}
(l_f C_{S, f}(g l_r-u_2 h_{c g}) x_3+(l_r C_{S, r}(g l_f\\
&+u_2 h_{c g})-l_f C_{S, f}(g l_r-u_2 h_{c g})) x_7-(l_f^2 C_{S, f}(g l_r \\
&-u_2 h_{c g})+l_r^2 C_{S, r}(g l_f+u_2 h_{c g})) \frac{x_6}{x_4}), \\
\dot{x}_7= & \frac{\mu}{x_4(l_r+l_f)}(C_{S, f}(g l_r-u_2 h_{c g}) x_3-(C_{S, r}(g l_f\\
&+u_2 h_{c g})+C_{S, f}(g l_r-u_2 h_{c g})) x_7+(C_{S, r}(g l_f \\
& +u_2 h_{c g}) l_r-C_{S, f}(g l_r-u_2 h_{c g}) l_f) \frac{x_6}{x_4})-x_6.
\end{aligned}
\end{equation}
where $f_{steer}$ and $f_{acc}$ impose physical constraints on steering and acceleration.
Readers are referred to CommonRoad documentation\footnote{\url{https://gitlab.lrz.de/tum-cps/commonroad-vehicle-models}} for the details of each parameter.

\begin{table}
\centering
\small
\caption{Adaptive cruise control (hyper)parameters.}
\label{table:acc_params}
\begin{tabular}{*{2}{c}}
\toprule
Parameter & Value\\
\midrule
Leading velocity $v_0$ & 14m/s\\
Desired velocity $v_d$ & 35m/s\\
Rock road friction coefficients $[f_1\; f_2\; f_3]$ & $[0.5\ 25\ 1.25]$\\
Rock road control bound coefficient $c_r$ & $0.5$\\
Dry road friction coefficients $[f_1\; f_2\; f_3]$ & $[0.3\ 15\ 0.75]$\\
Dry road control bound coefficient $c_d$ & $0.3$\\
Ice road friction coefficients $[f_1\; f_2\; f_3]$ & $[0.1\ 5\ 0.25]$\\
Ice road control bound coefficient $c_i$ & $0.1$\\
Look-ahead time $T_h$ & 1.8s\\
MPC horizon & 70 steps \\
MPC sampling frequency & 10Hz \\
Simulation sampling frequency & 100Hz\\
Simulation total time $T_{sim}$ & 20s\\
Neural network architecture & $[512\ 512\ 512]$ \\
Neural network activation & Sine \\
Training optimizer & Adam\\
Learning rates & $2\times10^{-5}, 8 \times 10^{-7}$ \\
Training epochs & 5K,55K,5K \\
Terminal time $T$ & 25s \\
\bottomrule
\end{tabular}
\end{table}

\begin{table}[h]
\centering
\small
\caption{Autonomous racing (hyper)parameters.}
\label{table:racing_params}
\begin{tabular}{*{2}{c}}
\toprule
Parameter & Value\\
\midrule
Friction coefficients $\mu_{low}, \mu_{mid}, \mu_{high}$ & $0.12, 0.6, 1.04$\\
Friction brake coefficient $\eta_{low}, \eta_{mid}, \eta_{high}$ & $0.01, 50, 500$\\
MPC horizon & 20 steps \\
MPC sampling frequency & 10Hz \\
Simulation sampling frequency & 100Hz\\
Neural network architecture & $[512\ 512\ 512]$ \\
Neural network activation & Sine\\ 
Training optimizer & Adam\\
Learning rates & $2\times10^{-5}, 8 \times 10^{-7}$
\\
Training epochs for left wall & 15K,165K,15K \\
Training epochs for right wall & 25K,250K,25K \\
Terminal time $T$ & 20s \\
\bottomrule
\end{tabular}
\end{table}

\end{document}